\documentclass[lettersize,journal]{IEEEtran}
\usepackage{amsmath,amsfonts}
\usepackage{algorithmic}
\usepackage{algorithm}
\usepackage{array}
\usepackage[caption=false,font=normalsize,labelfont=sf,textfont=sf]{subfig}
\usepackage{textcomp}
\usepackage{stfloats}
\usepackage{url}
\usepackage{verbatim}
\usepackage{graphicx}
\usepackage{cite}
\usepackage{amssymb}
\usepackage{multirow}
\usepackage[table,xcdraw]{xcolor}
\usepackage{threeparttable}
\usepackage{graphicx}
\usepackage{amsmath}
\usepackage{bm}
\usepackage{setspace}

\begin{document}

\title{Illumination-insensitive Binary Descriptor for Visual Measurement Based on Local Inter-patch Invariance}

\author{Xinyu Lin,~\IEEEmembership{Student Member,~IEEE}, 
			 Yingjie Zhou*,~\IEEEmembership{Member,~IEEE},  \\
			 Xun Zhang,~\IEEEmembership{Senior Member,~IEEE}, 				 
			 Yipeng Liu,~\IEEEmembership{Senior Member,~IEEE}, 	
	  and Ce Zhu*,~\IEEEmembership{Fellow,~IEEE}
	\newline
	\thanks{Xinyu Lin (xinyu.lin@std.uestc.edu.cn), Yipeng Liu, and Ce Zhu (eczhu@uestc.edu.cn) are with the Lab of Advanced Visual Communications \& Computing, School of Information and Communication Engineering, University of Electronic Science and Technology of China (UESTC), Chengdu 611731, China. Yingjie Zhou (yjzhou@scu.edu.cn) is with the College of Computer Science, Sichuan University, Chengdu, Sichuan, China. Xun Zhang is with the institut superieur d'electronique de Paris - ISEP,  Paris, France. (* Corresponding author: Ce Zhu and Yingjie Zhou.)
}}

\maketitle

\begin{abstract}
	Binary feature descriptors have been widely used in various visual measurement tasks, particularly those with limited computing resources and storage capacities. Existing binary descriptors may not perform well for long-term visual measurement tasks due to their sensitivity to illumination variations. It can be observed that when image illumination changes dramatically, the relative relationship among local patches mostly remains intact. Based on the observation, consequently, this study presents an illumination-insensitive binary (IIB) descriptor by leveraging the local inter-patch invariance exhibited in multiple spatial granularities to deal with unfavorable illumination variations. By taking advantage of integral images for local patch feature computation, a highly efficient IIB descriptor is achieved. It can encode scalable features in multiple spatial granularities, thus facilitating a computationally efficient hierarchical matching from coarse to fine. Moreover, the IIB descriptor can also apply to other types of image data, such as depth maps and semantic segmentation results, when available in some applications. Numerical experiments on both natural and synthetic datasets reveal that the proposed IIB descriptor outperforms state-of-the-art binary descriptors and some testing float descriptors. The proposed IIB descriptor has also been successfully employed in a demo system for long-term visual localization. The code\footnote{https://github.com/roylin1229/IIB\_descriptor} of the IIB descriptor will be publicly available.
\end{abstract}


\section{Introduction}
\label{sec_intro}
\IEEEPARstart{F}{eature} point descriptors could provide precise correspondences between different images of the same scenes. They lay the foundation for a wide range of visual measurement tasks, including localization \cite{9950428}, odometry \cite{9580965}, simultaneous localization and mapping (SLAM) \cite{9856705}, three-dimensional reconstruction \cite{8419781}, and servoing systems \cite{9751607}. In these tasks, feature points are first detected by using predefined operators \cite{SIFT, SURF, KAZE} or learning policies \cite{SuperPoint,ASLFeat,ALIKE}. Then, the neighborhood pixels, called point region of support (ROS), around feature points are used to form corresponding feature descriptors \cite{SIFT,SURF,KAZE,AKAZE,BEBLID,BRIEF,BRISK,BinBoost,FREAK,LATCH,LDB,LUCID,ORB}, which can be viewed as the signature of feature points. Feature descriptors should be compact, distinctive, and computationally efficient to facilitate processing, such as in feature matching \cite{gms-matching}. 

\begin{figure}[tbp]
	\centering
	\setlength{\abovecaptionskip}{0.cm}
	\setlength{\belowcaptionskip}{-0.cm}	
	\includegraphics[width=0.498\textwidth]{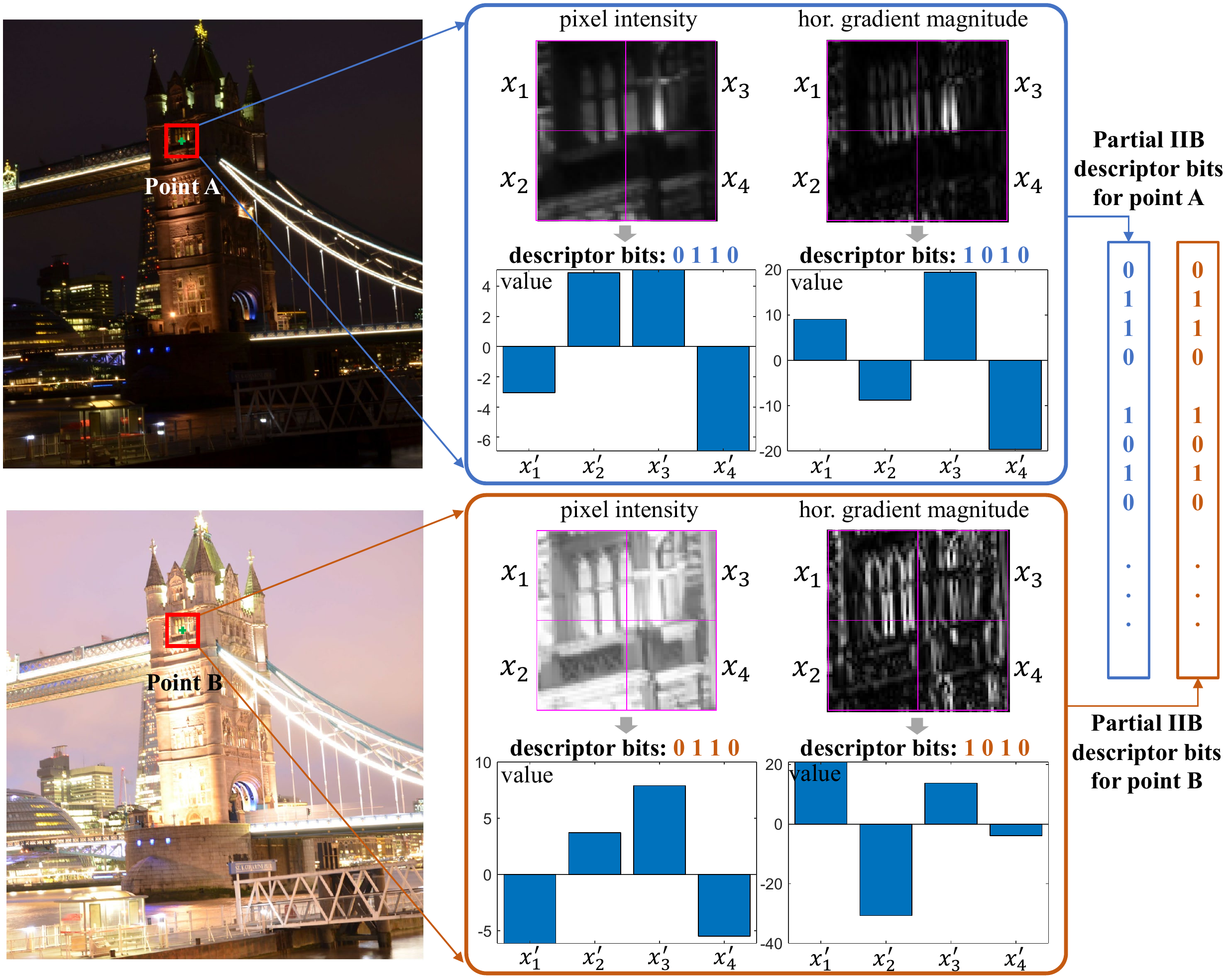}
	\caption{The example demonstrates the key observation in the IIB descriptor. When image illumination changes dramatically, the relative relationship among local patches mostly remains intact. Taking the channel for pixel intensity as an example, let $\bm{x}$ (denoted as $[x_1, x_2, x_3, x_4]$) be the average pixel intensity of the corresponding local patches. For point A and point B in the same location of pictures with different illumination conditions, although each value in $\bm{x}$ with respect to point B is larger than that associated with point A, the relative relationship regarding $x'_i$ (denoted as $x_i-mean(\bm{x}), i=1,2,3,4$) mostly remains intact, \textit{i.e.}, the corresponding descriptor bits for both points (A and B) are exactly the same as "0110" and "1010" for the channels of pixel intensity and horizontal gradient magnitude, respectively. Note that 0 represents that $x'_i$ has a value smaller than zero, while 1 represents that $x'_i$ has a value larger than zero.}
	\label{observation}
\end{figure}

Feature descriptors can be classified into float and binary descriptors. As the term suggests, float and binary descriptors are expressed using float and binary bit vectors, respectively. Compared with float descriptors \cite{9507474, 9761834, SIFT, SURF, KAZE, SuperPoint, ASLFeat, ALIKE}, binary descriptors \cite{AKAZE,BEBLID,BRIEF,BRISK,BinBoost,FREAK,LATCH,LDB,LUCID,ORB} have attracted increasing attention in visual measurement tasks because of their excellent storage and computation efficiency. They are more suitable for systems and platforms with limited computation and storage resources, such as mobile phones, drones, and robots. In addition, considering feature descriptor matching, descriptor similarity can be calculated using Hamming distance\footnote{https://en.wikipedia.org/wiki/Hamming\_distance} for binary descriptors, which is computationally more efficient than Euclidean distance\footnote{https://en.wikipedia.org/wiki/Euclidean\_distance} for float descriptors.  

\begin{figure}[tbp]
	\centering
	\setlength{\abovecaptionskip}{0.cm}
	\setlength{\belowcaptionskip}{-0.cm}
	\includegraphics[width=0.49\textwidth]{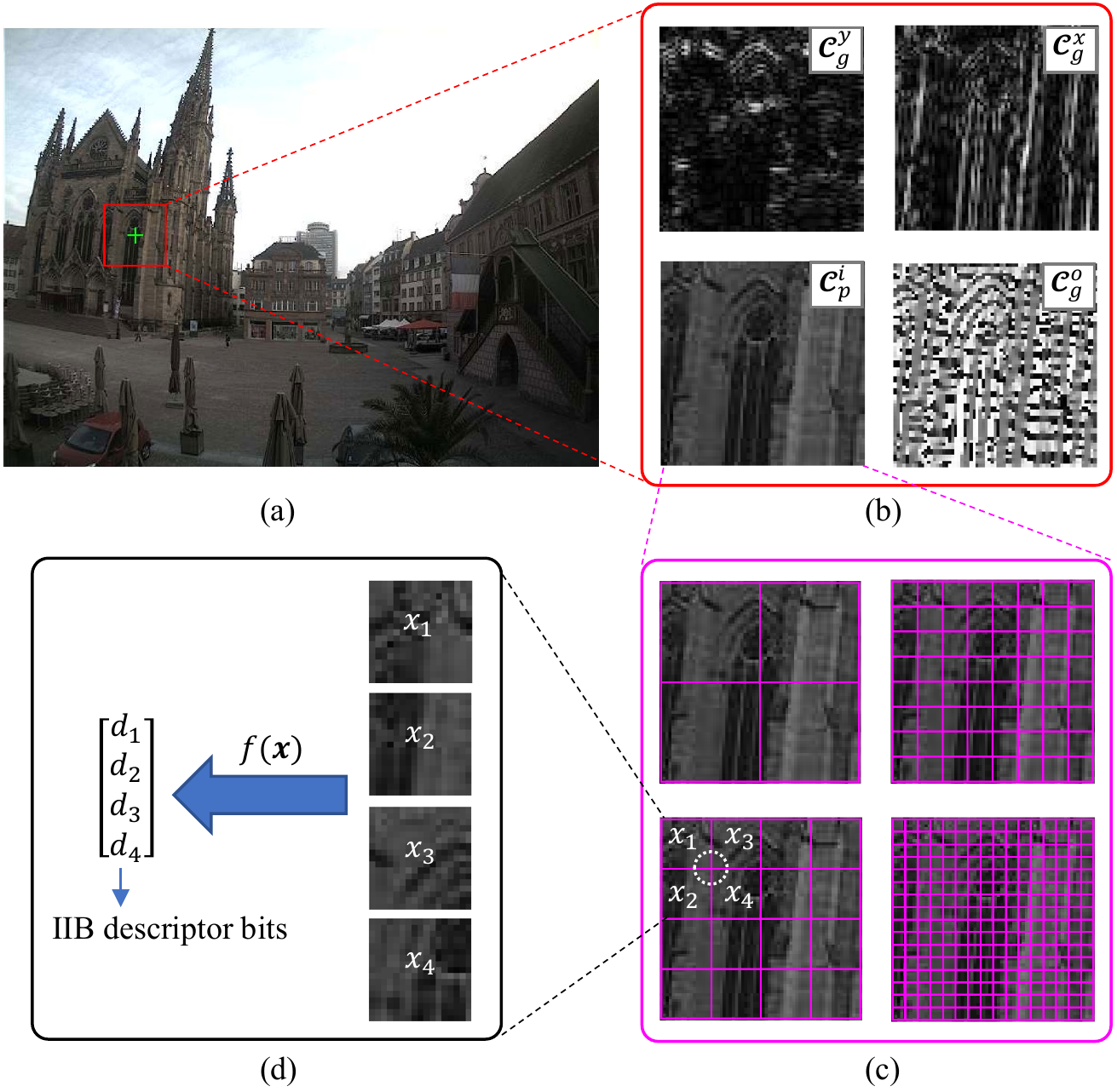}
	\caption{Formulation mechanism of the proposed IIB descriptor. (a) Feature point with corresponding ROS in a test image. (b) Multiple channels of image data (\textit{i.e.}, pixel intensity $\bm{\mathcal{C}}^i_p$, horizontal \& vertical gradient magnitudes $\bm{\mathcal{C}}^x_g$ \& $\bm{\mathcal{C}}^y_g$, and gradient orientation $\bm{\mathcal{C}}^o_g$). (c) Local patches split by a quadtree with multiple spatial granularities in feature point ROS. (d) Binary bit formulation based on the local inter-patch invariance. The IIB descriptor with binary bits $\bm{d}$ has long-term matching capability because the mapping function $f(\bm{x})$ on the quadruple of local patches can effectively exploit the relative illumination-insensitive relationship among them. The $x_i, i=1,2,3,4$ is the average image data value in the corresponding local patch, which can be calculated efficiently by integral images.}
	\label{fig_demo_v2}
\end{figure}

Binary descriptors use fewer bits to encode feature points, indicating that they lose some information, and more truncation errors are introduced. Generally, in terms of performance, they are mediocre compared with float descriptors, particularly when image illumination changes considerably, which is a general case of long-term visual measurement tasks. As mentioned in \cite{LongTermVisualLocalizationRevisited}, problems still exist in illumination-insensitive image feature detection, description, and matching for many long-term visual measurement tasks. 

\textbf{\textit{This study proposes an illumination-insensitive binary (IIB) descriptor based on the local inter-patch invariance presented in multiple spatial granularities to address the feature description and matching problems under drastic illumination variations}}. Fig. \ref{fig_demo_v2} depicts the formulation mechanism of the proposed IIB descriptor. As shown in Fig. \ref{fig_bit_rules}, existing binary descriptors generally use pairs of sampled points \cite{BRIEF,BRISK,ORB} or sampled local patches \cite{BEBLID,LDB,AKAZE} in feature point ROS to form descriptors. Instead, the quadruple inter-patch invariance of all local patches in feature point ROS is used to formulate the IIB descriptor. As shown in Fig. \ref{observation}, when image illumination changes drastically, the image appears brighter or darker entirely. However, the relative relationship among local patches in feature point ROS mostly remains intact. The relative relationship among local patches can be exploited by analyzing their local inter-patch invariance, which is critical for the illumination-insensitive ability of the IIB descriptor.

The local inter-patch invariance is collected hierarchically in multiple spatial granularities to encode feature points from coarse to fine, as shown in Fig. \ref{fig_demo_v2} (c). \textbf{\textit{This formulation mechanism can be used to perform computationally efficient hierarchical matching from coarse to fine instead of brute-force matching}} by assuming that two IIB descriptors are similar only if their binary bits in each granularity are similar.

\textbf{\textit{The proposed IIB descriptor is scalable}}. By default, pixel intensity and gradients (\textit{i.e.}, gradient orientation and horizontal \& vertical gradient magnitudes) of gray images serve as channels for the IIB descriptor formulation. As shown in Fig \ref{fig_channels}, other types of image data, which are essentially illumination-insensitive, can be used as additional channels to enhance the performance of the IIB descriptor if available in specific visual measurement tasks. Examples of these image data could be depth, texture \cite{LBP}, and semantic segmentation maps from other sensors and algorithms.

\begin{figure}[tbp]
	\centering
	\setlength{\abovecaptionskip}{0.cm}
	\setlength{\belowcaptionskip}{-0.cm}	
	\includegraphics[width=0.49\textwidth]{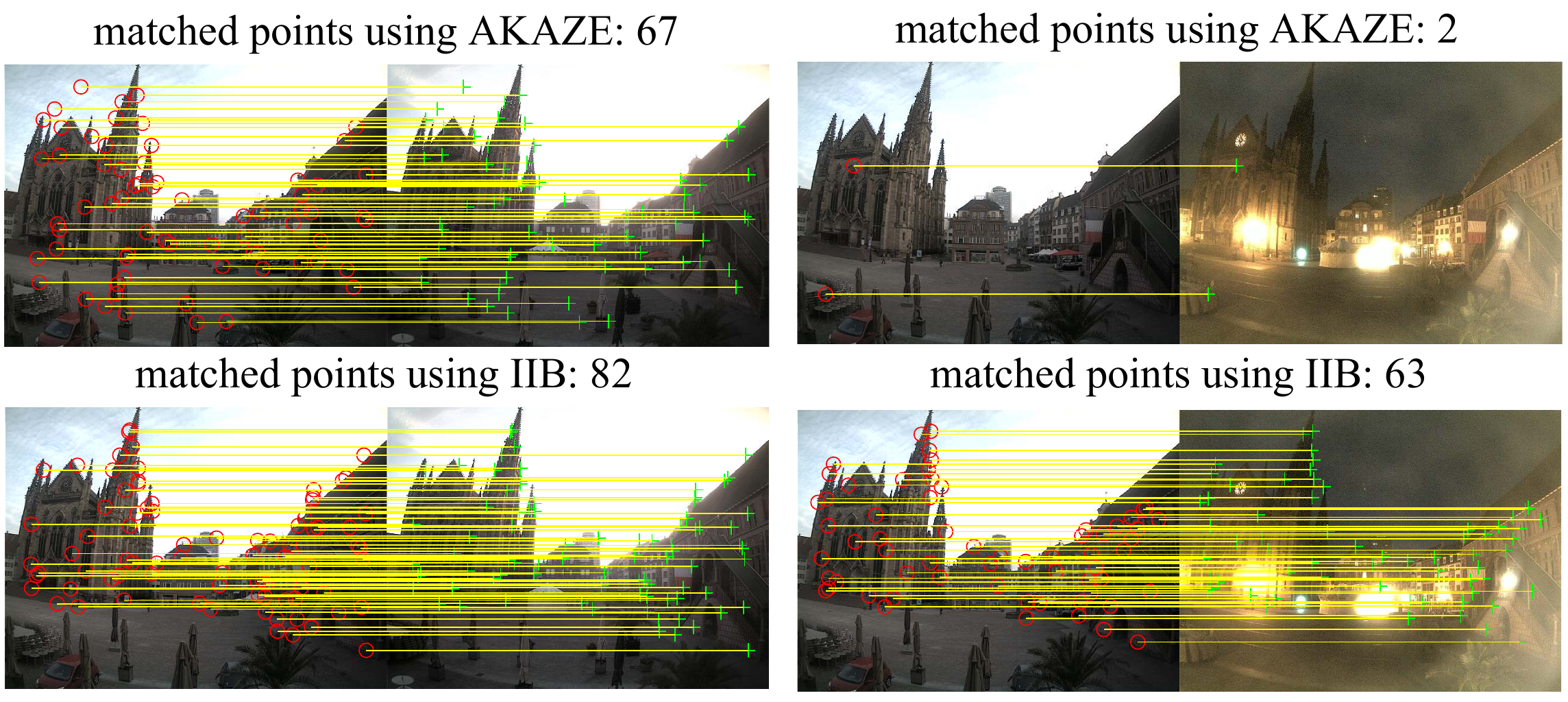}
	\caption{Examples of matched points using mutual brute-force matching of the AKAZE \cite{AKAZE} binary descriptor (top row) and proposed IIB descriptor (bottom row) for test images with similar (first column) and poles apart (second column) illumination conditions. Fig. \ref{visual_performance} shows more visual comparisons.}
	\label{fig_demo}
\end{figure}

Numerical experiments on both natural and synthetic datasets show that the proposed IIB descriptor outperforms other state-of-the-art (SOTA) binary descriptors and some testing float descriptors. Fig. \ref{fig_demo} shows visualization examples of matched points using the AKAZE \cite{AKAZE} and IIB descriptors. The corresponding fast hierarchical matching policy can significantly reduce matching costs compared with the brute-force matching policy while maintaining competitive performance. Using additional channels of image data can further improve the performance of the IIB descriptor if available in specific visual measurement tasks. \textbf{\textit{The IIB descriptor is highly efficient because of the usage of integral images \cite{integralimage}, particularly when the rotation of point features is not considered}}, which is a general case for many visual measurement tasks, \textit{e.g.,} visual SLAM for robots. 

The contributions of this paper are summarized as follows.
\begin{enumerate}
	\item This study proposes an illumination-insensitive binary descriptor by leveraging the local inter-patch invariance exhibited in multiple spatial granularities to deal with unfavorable illumination variations. It is based on the observation that when image illumination changes dramatically, the relative relationship among local patches mostly remains intact.
	\item The IIB descriptor can encode feature points in multiple spatial granularities, thus facilitating a computationally efficient hierarchical matching from coarse to fine.
	\item The IIB descriptor is scalable and can be extended using additional image data if available.
	\item Numerical experiments reveal the superiority of the IIB descriptor over other state-of-the-art descriptors in terms of both extraction efficiency and performance.
\end{enumerate}

The remainder of this paper is organized as follows. Section \ref{sec_related_works} introduces related work, particularly those SOTA binary descriptors. Section \ref{sec_IIB} describes the proposed IIB descriptor in detail. Section \ref{sec_exp} presents numerical experiments and related discussions, where the proposed IIB descriptor is compared with some SOTA binary and float descriptors from multiple aspects using natural and synthetic datasets. Section \ref{sec_app} presents a demo system for long-term visual localization in visual measurement using the proposed IIB descriptor. Finally, Section \ref{sec_conclusion} summarizes the conclusion.

\section{Related Work}
\label{sec_related_works}
As mentioned in Section \ref{sec_intro}, many binary descriptors have been proposed over the past two decades. A comprehensive review of these studies is beyond the scope of this study. Here, only related SOTA representations, particularly those integrated into OpenCV\footnote{https://opencv.org/} and used for the comparison in this study, are briefly reviewed. Please refer to  \cite{ComparativeEvaluationofBinaryFeatures,AnExperimentalEvaluationofBinaryFeatureDescriptors,7583687} for more reviews of binary descriptors.

\begin{figure}[tbp]
	\centering
	\setlength{\abovecaptionskip}{0.cm}
	\setlength{\belowcaptionskip}{-0.cm}	
	\includegraphics[width=0.49\textwidth]{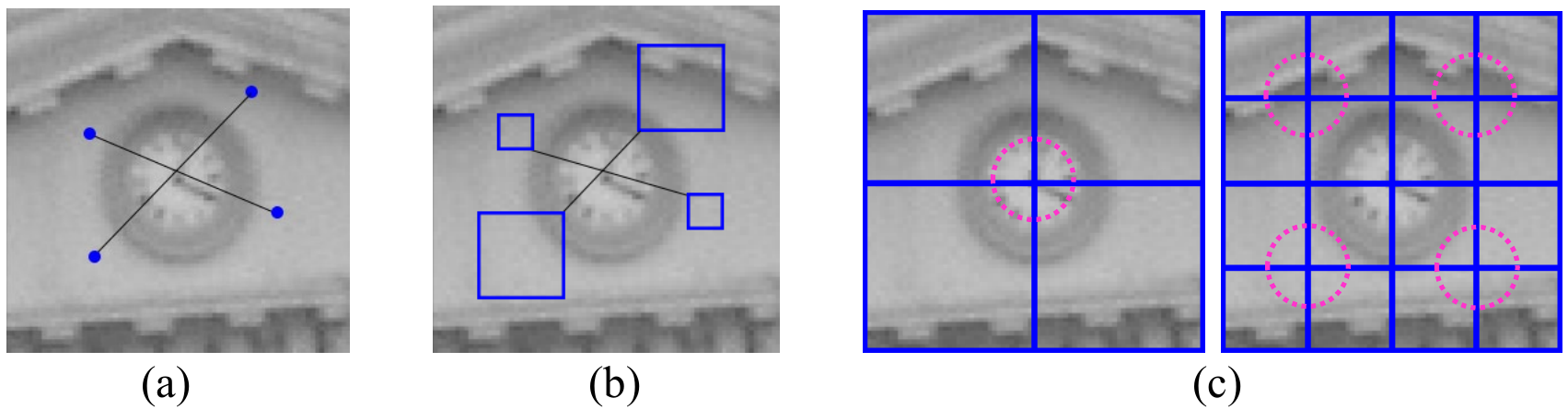}
	\caption{(a) Two pairs of sampled points in the BRIEF descriptor; (b) Two pairs of sampled local patches in the BEBLID descriptor; (c) Local patches at the first and second granularities in the proposed IIB descriptor. Compared with the first two descriptors using only sampled image data in feature point ROS, the proposed approach leverages all image data in corresponding ROS. In addition, instead of using pairs of sampled points and local patches shown in (a) and (b), quadruples of local patches are used in the proposed approach to better encode feature points.} 	
	\label{fig_bit_rules}
\end{figure}

The binary robust independent elementary feature (BRIEF) \cite{BRIEF} is the most representative binary descriptor. The following binary descriptors are based on it, to an extent. As shown in Fig. \ref{fig_bit_rules} (a), pairs of sampled points in feature point ROS obeying a particular distribution around feature points are first selected in the BRIEF descriptor. The intensities of these pairs of sampled points are then compared to formulate binary descriptors. Because the BRIEF descriptor is not rotation-invariant, Rublee et al. proposed the steered BRIEF descriptor \cite{ORB} based on the orientation in feature point ROS.  With orientated features from accelerated segment test (FAST) \cite{FAST}, the combined feature is called oriented FAST and rotated BRIEF (ORB) \cite{ORB} feature, which is widely used in visual measurement systems,  \textit{e.g.}, visual SLAM \cite{ORB-SLAM3} systems. In binary robust invariant scalable keypoints (BRISK) \cite{BRISK}, sampled points lying on scaled concentric circles of feature points are selected to construct the BRISK descriptor, exhibiting remarkable performances on the benchmark \cite{vggaffine}. Alahi et al. proposed a fast retina keypoint (FREAK) \cite{FREAK} descriptor, which was inspired by the human visual system and formulated based on retinal sampled points. 

Generally, to make descriptors insensitive to noise, the information on sampled points is smoothed with a Gaussian kernel \cite{BRIEF,BRISK,ORB,FREAK}, which can improve the robustness of descriptors to an extent. In \cite{LDB}, the authors proposed a local difference binary (LDB) descriptor, in which pairs of sampled patches in feature point ROS instead of sampled points are used to construct a binary descriptor. With the help of an integral image \cite{integralimage}, the computational efficiency is remarkable compared with other descriptors, particularly when features are upright. Alcantarilla et al. proposed a modified LDB descriptor for the AKAZE feature \cite{AKAZE}, in which sampled patches are sub-sampled again according to the scale of features. Except for pixel intensity, orientated local gradients at multiple spatial granularities are used to construct the descriptors in the LDB \cite{LDB} and AKAZE \cite{AKAZE}. In boosted efficient binary local image descriptor (BEBLID) \cite{BEBLID}, a learned generation scheme for pairs of sampled patches was introduced. Fig. \ref{fig_bit_rules} (b) shows an example of sampled patch pairs for the BEBLID descriptor.

The sampled points mentioned in \cite{BRIEF,BRISK,ORB,FREAK} are generated according to handcrafted rules, such as a Gaussian or uniform distribution. Some studies generated sampled points/patches based on learning policies \cite{BinBoost,LATCH,BEBLID}. Trzcinski et al. proposed BOOST \cite{BinBoost}, which is computed using a boosted binary hash function. Instead of learning pairs of sampled points, the descriptor based on learned arrangements of three patch codes (LATCH) was proposed in \cite{LATCH}. In \cite{BEBLID}, the authors used BoostedSSC with an improved weak-learner training strategy to form the BEBLID descriptor. Similarly, AdaBoost has been used in the LDB descriptor \cite{LDB} and the AKAZE feature \cite{AKAZE} to sample local patches.

Other studies have not depended on sampled points or patches. In \cite{LUCID}, the authors proposed a descriptor based on permutation distances of the ordering of RGB values. Some deeply learned binary descriptors \cite{L2-Net,bandara2021deep,9206151,9169844} have recently been proposed. Although some of them outperform SOTA handcrafted binary descriptors on their evaluation datasets, they are not widely applied in visual measurement tasks because of their high computational cost. Many of these methods are hard to achieve real time feature extraction, even with the help of GPU devices.

Compared to existing SOTA binary descriptors, \textit{e.g.}, the AKAZE \cite{AKAZE}, the IIB descriptor presents some improvements. They are as follows: (1) instead of using pairs of sampled local patches, all local patches split by a quadtree in multiple spatial granularities of feature point ROS are used to construct the IIB descriptor; (2) unlike using the binary comparison based on pairs of sampled local patches, the local inter-patch invariance is analyzed among every four local patches, which exploits their relative relationship to reach the illumination-insensitive characteristic of the IIB descriptor; and (3) the IIB descriptor can be extended using more channels of image data, if available in specific visual measurement tasks, and can be matched using computationally efficient hierarchical matching instead of brute-force matching.

\section{IIB Descriptor}
\label{sec_IIB}
As shown in Fig. \ref{fig_demo_v2}, the two critical formulation steps of the IIB descriptor can be summarized as follows: (1) local patch generation in feature point ROS and (2) binary bit formulation according to the local patches generated. The local patch generation mechanism can be further analyzed to perform fast hierarchical matching instead of brute-force matching. Binary bit formulation can be performed using multiple channels of image data and extended using additional image data, such as depth maps for visual measurements. Algorithm \ref{alg1} shows a procedure for the IIB descriptor formulation.

\begin{algorithm}[tbp]
	\caption{The IIB descriptor formulation procedure.}\label{alg:IIB}
	\hspace*{0.06in}\textbf{Input:} \\
	\hspace*{0.5cm} A test image $\bm{\mathcal{I}} \in \mathcal{R}^{W\times H}$ \\
	\hspace*{0.5cm} Detected feature points $\bm{p}^k \in \mathcal{R}^{4\times 1}, k=1,2,...,K$ in $\bm{\mathcal{I}}$ \\
	\hspace*{0.06in}\textbf{Output:} \\
	\hspace*{0.5cm} Extracted IIB descriptors $\bm{d}^k \in \{0, 1\}^{M\times 1}, k=1,2,...,K$\\
	\begin{algorithmic}[1]
		\STATE Obtain image data $\bm{\mathcal{C}} \in \mathcal{R}^{W\times H}$ in $N$ different channels, \textit{e.g.}, $\bm{\mathcal{C}}^i_p$ shown in Fig. \ref{fig_demo_v2}, based on the test image $\bm{\mathcal{I}}$;
		\FOR{$k=1$ to $K$}
			\STATE Obtain the ROS for feature point $\bm{p}^k$;
			\STATE Generate local patches in multi-spatial granularities $g=1,2,...,G$ from the point ROS (Subsection \ref{sec_IIB_sampling});
			\FOR{local patches in each granularity $g$}
				\FOR{image data $\bm{\mathcal{C}}$ in each channel}
					\STATE Calculate the IIB descriptor bits ${d^k_j}$ according to mapping functions $f(\bm{x})$ based on image data $\bm{\mathcal{C}}$ (Subsection \ref{sec_IIB_bit});
				\ENDFOR
			\ENDFOR
			\STATE Concatenate a series of the IIB descriptor bits $d^k_j, j = 1,2,...,M$ obtained in $N$ image channels and $G$ spatial granularities as the final descriptor vector $\bm{d}^k$; 
			\RETURN  $\bm{d}^k$
		\ENDFOR
	\end{algorithmic}
	\hspace*{0.5cm} \\
	Note: $\bm{d}^k$ can be matched hierarchically (Subsection \ref{sec_hierarchical_matching}) and sampled optionally (Subsection \ref{sec_adaboost_sampling}.)
	\label{alg1}
\end{algorithm}

\subsection{Local Patch Generation}
\label{sec_IIB_sampling}
The local patches of the IIB descriptor are generated according to a recursive quadtree split in feature point ROS, which may be rotated according to feature point orientations if they are considered. In particular, each local patch in the current granularity is equally split into four local mini-patches in the next granularity with more fine-grained features. Fig. \ref{fig_demo_v2} (c) and  Fig. \ref{fig_bit_rules} (c) show examples in which the local patches of the first granularity are generated by equally splitting the feature point ROS, and four mini-patches for each local patch are generated by splitting it again into the second granularity. Local patches with a larger granularity can be created in the same manner. The number of local patches at each granularity is $4^g, g = 1,2,3,..., G$, where $g$ denotes the level of granularity and, $G$ is the maximal level of granularity. 

Using the process described above, local patches with multiple granularities correspond to different local patch sizes. The ones with large patch sizes (coarse granularity) could represent more general features by using fewer descriptor bits, while the ones with small patch sizes (fine granularity) consider more detailed features by employing more descriptor bits. Local patches at multiple granularities can be used to encode feature points from coarse to fine, which is critical for the fast hierarchical matching described in Subsection \ref{sec_hierarchical_matching}.

\subsection{Binary Bit Formulation}
\label{sec_IIB_bit}
\begin{figure}[tbp]
	\centering
	\setlength{\abovecaptionskip}{0.cm}
	\setlength{\belowcaptionskip}{-0.cm}	
	\includegraphics[width=0.49\textwidth]{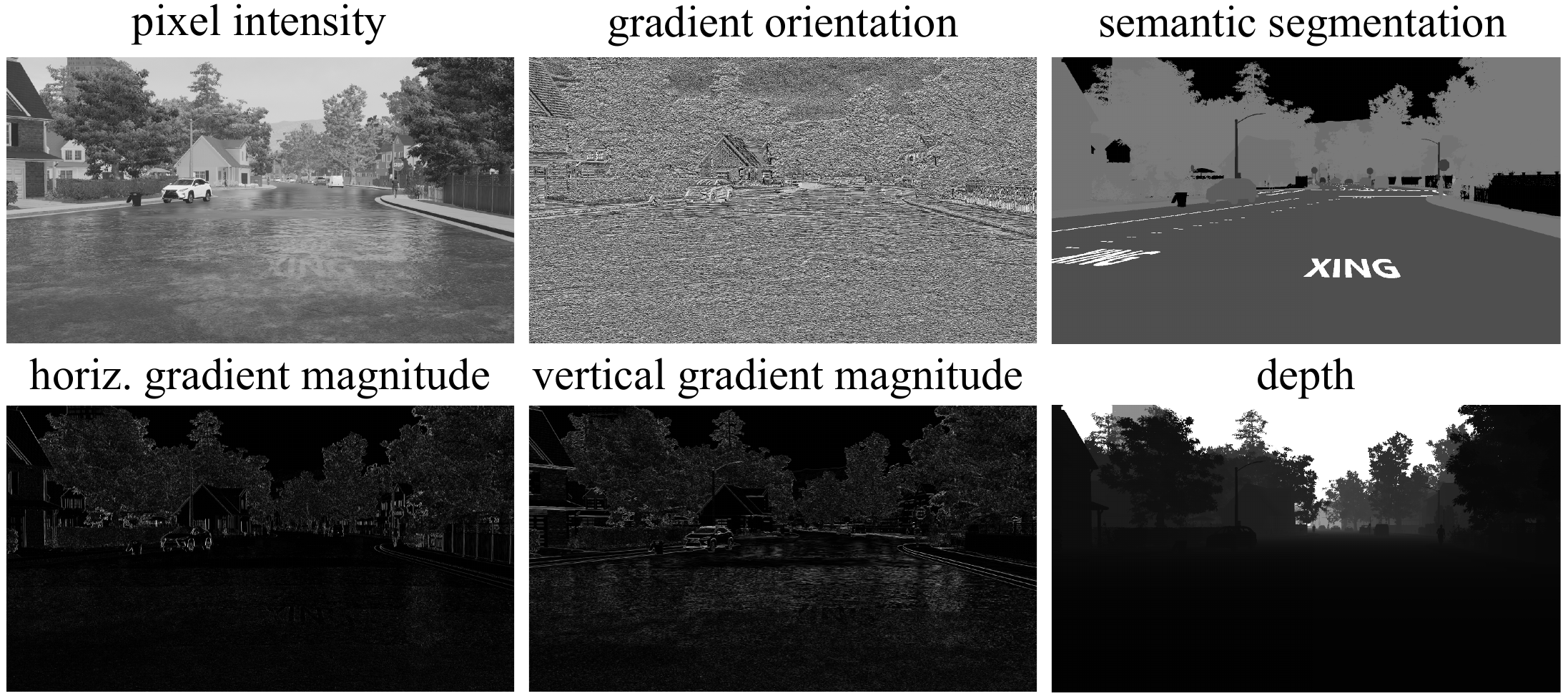}
	\caption{Image data for the IIB descriptor formulation, in which the channels of pixel intensity and related gradients (\textit{i.e.}, gradient orientation and horizontal \& vertical gradient magnitudes) are default, as in \cite{IILB}, and semantic segmentation and depth maps are optional.}
	\label{fig_channels}
\end{figure}

Existing binary descriptors generally use pairs of sampled points or local patches in feature point ROS to construct descriptors. In contrast, the inter-patch invariance among every four local patches exhibited in multiple spatial granularities of feature point ROS is used to formulate the IIB descriptor. In particular, as shown in Fig. \ref{fig_demo_v2} (c) and (d), four nonoverlapped local patches in the neighborhood are used to calculate the IIB descriptor bits according to the mapping function $f(\bm{x})$, where $\bm{x}$ (denoted as $[x_1, x_2, x_3, x_4]$) is the average value of pixel intensity or gradient features in corresponding local patches. As shown in Fig. \ref{fig_channels}, $\bm{x}$ can also be the average value of additional image data, such as from semantic segmentation results and depth maps for visual measurements. Using four overlapped local patches can theoretically improve the IIB descriptor performance to an extent. However, the descriptor size would be significantly increased. 

The mapping function $f(\bm{x})$ fully exploits the relative relationship among local patches in the neighborhood to make the descriptor robust to illumination variations. It is based on the assumption that when image illumination changes drastically, the relative relationship among local patches in feature point ROS mostly remains intact. Five mapping functions $f(\bm{x})$, in which $\bm{x}$ is computed using integral images \cite{integralimage}, were introduced and compared experimentally. They can be described as follows, where $\bm{x}$ is the vector of $x_i, i = 1,2,3,4$.

\subsubsection{Mean Mapping} 
\begin{equation}
	d_i = \begin{cases}
		1, & x_i > mean(\bm{x}) \\
		0, & otherwise
	\end{cases}
\end{equation}
This straightforward and effective mapping rule is recommended/default mapping function. The mean mapping of pixel intensity can directly indicate which local patches in the neighborhood are brighter or darker. It can also report which local patches in the neighborhood include more structured features using image gradients. In addition, $mean(\bm{x})$ can be obtained directly from the previous granularity, and computing them again is unnecessary.

\subsubsection{Max/Min Mapping}
Similar to the mean mapping, in max/min mappings, the $x_i > mean(\bm{x})$ for mean mapping is replaced by $x_i = max(\bm{x})$ for $max$ mapping and $x_i = min(\bm{x})$ for $min$ mapping.

\subsubsection{Quartile Mapping}
In quartile mapping, let $R = max(\bm{x})-min(\bm{x})$. Then, 
\begin{equation}
	d_i = \begin{cases}
		11, & x_i - min(\bm{x}) > 0.75R    							\\
		10, & 0.5R < x_i - min(\bm{x}) <= 0.75R  				\\
		01, & 0.25R < x_i - min(\bm{x}) <= 0.5R				\\
		00, & Otherwise									\\
	\end{cases}
\end{equation}
The quartile mapping can be viewed as an extension of the mean mapping as it uses two bits to encode $d_i$.

\subsubsection{Sort Mapping}
In sort mapping, $\bm{x}$ is sorted first. Then, similar to quartile mapping, two bits are used to encode $d_i$ according to their order of $x_i$.

\subsection{Fast Hierarchical Matching for the IIB Descriptor}
\label{sec_hierarchical_matching} 
According to the IIB descriptor formulation mechanism, binary bits increase exponentially with increasing granularity. The IIB descriptor size is $N\times\sum_{g=1}^{G}4^g$, where $G$ is the maximal level of granularity, and $N$ is the number of image data channels. Performing brute-force matching of two sets of the IIB descriptors involves massive bit comparisons. Because the IIB descriptor is constructed in multiple spatial granularities of feature point ROS, which encodes feature points from coarse to fine, two IIB descriptors are similar only if their binary bits in each granularity are similar. The binary bits of the IIB descriptor with low granularity can be first compared to reduce matching costs and eliminate many dissimilar descriptors. The remaining parts use binary bits with high granularity to perform precise matching. In particular, only descriptor pairs with a Hamming distance smaller than a certain percentage of bit numbers in current granularity are further delivered to the following granularity for comparison.

\subsection{Reduction of the IIB Descriptor Size (Optional)}
\label{sec_adaboost_sampling} 
As aforementioned, the IIB descriptor can be customized in multiple spatial granularities and image data channels. The size of the IIB descriptor changes according to these parameters. Therefore, an intuitive manner of reducing the IIB descriptor size is to choose an economic parameter combination scheme, such as using less granularity and image data channels. 

Inspired by other SOTA methods \cite{AKAZE,BEBLID,LDB}, another alternative is to perform sampling for the quadruple of local patches (shown in Fig. \ref{fig_bit_rules} (c)) based on machine learning techniques instead of using all of them. Here, Adaboost \cite{FREUND1997119} was used to learn the weight of these quadruples, which can be further analyzed in the sampling process. The quadruples of local patches with top-$M$ weights, where $M$ is the desired size, were used to calculate the IIB descriptor.

\section{Numerical Experiments}
\label{sec_exp}
The performance of the proposed IIB descriptor was thoroughly evaluated on both natural and synthetic datasets from multiple aspects to demonstrate its superiority over popular descriptors, which are widely used in many visual measurement tasks. In particular, various SOTA OpenCV integrated feature descriptors were compared with the proposed IIB descriptor because these descriptors have been widely verified and validated both in academia and industry for visual measurement applications. These compared descriptors were BEBLID \cite{BEBLID}, ORB \cite{ORB}, BRISK \cite{BRISK}, BOOST \cite{BinBoost}, BRIEF \cite{BRIEF}, LATCH \cite{LATCH}, AKAZE \cite{AKAZE}, FREAK \cite{FREAK}, SIFT \cite{SIFT}, SURF \cite{SURF}, and KAZE \cite{KAZE}. In addition, some deep learning-based methods, including a binary descriptor called L2-Net \cite{L2-Net} and three float descriptors, SUPERPOINT \cite{SuperPoint}, ASLFEAT \cite{ASLFeat}, and ALIKE \cite{ALIKE}, were also considered in numerical experiments to demonstrate the superior performance of the IIB descriptor. 

\subsection{Datasets and Metrics} 
\label{eva_data_metric}
Comparisons were conducted on both natural and synthetic datasets. The natural dataset comprised 906 test images with illumination variations extracted from the VggAffine \cite{vggaffine}, HPatches \cite{HPatches}, and Webcam \cite{TILDE} datasets. The synthetic dataset comprised 2087 test images with illumination variations selected from the Apollo Synthetic dataset\footnote{https://apollo.auto/synthetic.html}. Furthermore, related depth and semantic segmentation images were provided in the synthetic dataset, which was used to further validate the scalability of the IIB descriptor. The homography matrices for all testing image pairs were available to provide point-to-point correspondences, as described in \cite{vggaffine}.

The $precision$ and $recall$ defined in the benchmark \cite{vggaffine} were adopted as the principal criteria, which are widely used for evaluation purposes in visual measurement tasks. They were formulated as follows 
\begin{equation}
	precision = \frac{\# \text{\textit{correct matches}}}{\# \text{\textit{putative matches}}}, 
\end{equation}
and
\begin{equation}
	recall = \frac{\# \text{\textit{correct matches}}}{\# \text{\textit{correspondences}}}. 
\end{equation}
The two descriptors are considered a \textit{putative match} only when they are the mutual best match for each other using brute-force matching. A \textit{putative match} is considered a \textit{correct match} only when the corresponding two points satisfy geometry validation according to the homography matrix \cite{vggaffine}. The definition of \textit{correspondence} is given in \cite{vggaffine}. 

Both feature descriptors and detectors influence the $precision$ and $recall$ metrics. For all experiments, the parameters of feature detectors were the same for compared descriptors to eliminate the influence of feature detectors and only consider the performance of feature descriptors.

In addition to $precision$ and $recall$ metrics, to evaluate fast hierarchical matching for the IIB descriptor, the match cost was defined as $ MC = N_{hierarchical} / N_{brute-force}$, where $N_{hierarchical}$ and $N_{brute-force}$ are the numbers of descriptor bit comparisons of fast hierarchical matching and brute-force matching strategies, respectively.

\subsection{Parameter Settings of the IIB Descriptor}
According to the IIB descriptor mechanism, some critical parameters significantly affect the performance of the IIB descriptor. They were determined based on the experimental configuration described in Subsection \ref{sec_exp_A}.

\subsubsection{Choice of Image Data Channels} 
As mentioned in Subsection \ref{sec_IIB_bit}, multiple image data channels can be used to construct the IIB descriptor. TABLE \ref{IIB_with_channels} lists the \textit{mean average precision} (\textit{mAP}), \textit{mean average recall} (\textit{mAR}), and descriptor sizes of the IIB descriptors with different image data channels. When considering single channels, the $\bm{\mathcal{C}}^x_g$ or $\bm{\mathcal{C}}^y_g$ channels outperform the $\bm{\mathcal{C}}^o_g$ and $\bm{\mathcal{C}}^i_p$ channels. Both pixel intensity ($\bm{\mathcal{C}}^i_p$) and related gradients (\textit{i.e.}, gradient orientation $\bm{\mathcal{C}}^o_g$ and horizontal \& vertical gradient magnitudes $\bm{\mathcal{C}}^x_g$ \& $\bm{\mathcal{C}}^y_g$ in TABLE \ref{IIB_with_channels}) were used to construct the IIB descriptor by default. 

\begin{figure}[tbp]
	\centering
	\setlength{\abovecaptionskip}{0.cm}
	\setlength{\belowcaptionskip}{-0.cm}	
	\includegraphics[width=0.49\textwidth]{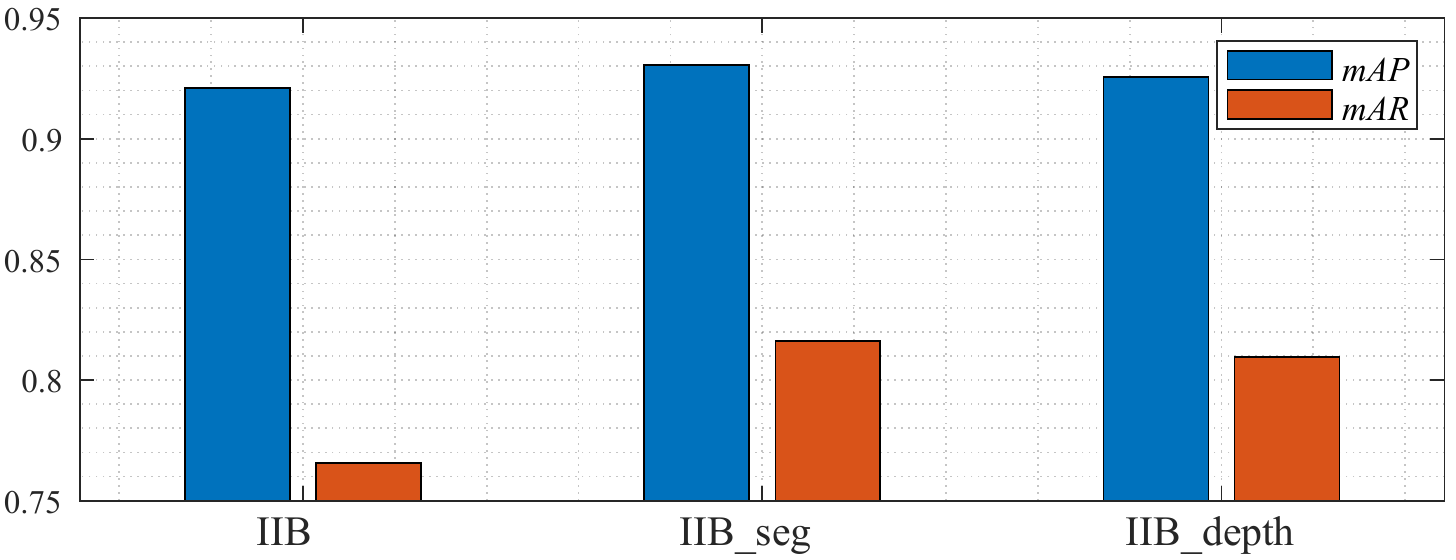}
	\caption{Performance comparison of the IIB descriptors with additional image data, in which the level of granularity is 4. The results show that the IIB descriptor is scalable.}
	\label{fig_additional_image_info}
\end{figure}

The proposed IIB descriptor is scalable. More illumination-insensitive image data can be used to further improve the IIB performance. Fig. \ref{fig_additional_image_info} shows two examples of this, in which two IIB descriptors extended using depth and semantic segmentation data are compared with the standard IIB descriptor. Evidently, using additional image data can improve the performance of the IIB descriptor, particularly for the $recall$ metric. 

\begin{table}[tbp]
	\centering
	\scriptsize
	\begin{threeparttable}[b]
		\caption{\textit{mAP},  \textit{mAR}, and descriptor size of the IIB descriptors with different image data channels, in which the $\bm{\mathcal{C}}^x_g$, $\bm{\mathcal{C}}^y_g$, $\bm{\mathcal{C}}^o_g$, and $\bm{\mathcal{C}}^i_p$ denotes the horizontal gradient magnitude, vertical gradient magnitude, gradient orientation, and pixel intensity, respectively.}
		\label{IIB_with_channels}
		\begin{tabular}{|c|cc|cc|c|}
			\hline
			\multirow{2}{*}{Descriptor} & \multicolumn{2}{c|}{Natural dataset}                   & \multicolumn{2}{c|}{Synthetic dataset}                 & \multirow{2}{*}{\begin{tabular}[c]{@{}c@{}}Descriptor \\ size (bits)\end{tabular}} \\ \cline{2-5}
			& \multicolumn{1}{c|}{\textit{mAP}}             & \textit{mAR}             & \multicolumn{1}{c|}{\textit{mAP}}             & \textit{mAR}             &                                                                                     \\ \hline
			{IIB-$\bm{\mathcal{C}}^y_g$}                           & \multicolumn{1}{c|}{0.8282}          & 0.6568          & \multicolumn{1}{c|}{0.8365}          & 0.6382          & 340                                                                                 \\ \hline
			{IIB-$\bm{\mathcal{C}}^x_g$}                           & \multicolumn{1}{c|}{0.8391}          & 0.6678          & \multicolumn{1}{c|}{0.8318}          & 0.6365          & 340                                                                                 \\ \hline
			{IIB-$\bm{\mathcal{C}}^i_p$}                           & \multicolumn{1}{c|}{0.8303}          & 0.6710          & \multicolumn{1}{c|}{0.7818}          & 0.5709          & 340                                                                                 \\ \hline
			{IIB-$\bm{\mathcal{C}}^o_g$}                           & \multicolumn{1}{c|}{0.7417}          & 0.5896          & \multicolumn{1}{c|}{0.7570}          & 0.5681          & 340                                                                                 \\ \hline
			{IIB-($\bm{\mathcal{C}}^o_g$+$\bm{\mathcal{C}}^i_p$)}                        & \multicolumn{1}{c|}{0.8596}          & 0.7232          & \multicolumn{1}{c|}{0.8425}          & 0.6553          & 680                                                                                 \\ \hline
			{IIB-($\bm{\mathcal{C}}^x_g$+$\bm{\mathcal{C}}^y_g$)}                        & \multicolumn{1}{c|}{0.9012}          & 0.7515          & \multicolumn{1}{c|}{0.9008}          & 0.7275          & 680                                                                                 \\ \hline
			{IIB-($\bm{\mathcal{C}}^x_g$+$\bm{\mathcal{C}}^y_g$+$\bm{\mathcal{C}}^i_p$)}                     & \multicolumn{1}{c|}{0.9134}          & 0.7918          & \multicolumn{1}{c|}{0.9129}          & 0.7465          & 1020                                                                                \\ \hline
			{IIB-($\bm{\mathcal{C}}^x_g$+$\bm{\mathcal{C}}^y_g$+$\bm{\mathcal{C}}^o_g$)}                     & \multicolumn{1}{c|}{0.9181}          & 0.7863          & \multicolumn{1}{c|}{0.9188}          & 0.7644          & 1020                                                                                \\ \hline
			{IIB-($\bm{\mathcal{C}}^x_g$+$\bm{\mathcal{C}}^y_g$+$\bm{\mathcal{C}}^o_g$+$\bm{\mathcal{C}}^i_p$)}                  & \multicolumn{1}{c|}{\textbf{0.9187}} & \textbf{0.8041} & \multicolumn{1}{c|}{\textbf{0.9210}} & \textbf{0.7655} & 1360                                                                                \\ \hline
		\end{tabular}
		\begin{tablenotes}
			\item The level of granularity is 4.
		\end{tablenotes}
	\end{threeparttable}
\end{table}

\subsubsection{Choice of Mapping Function $f(\textbf{x})$}
As mentioned in Subsection \ref{sec_IIB_bit}, in the framework of the IIB descriptor, different mapping functions $f(\bm{x})$ can formulate diverse IIB descriptors. TABLE \ref{IIB_with_mapping_rules} lists the \textit{mAP}, \textit{mAR}, and descriptor size of the IIB descriptors with five mapping functions $f(\bm{x})$ mentioned in Subsection \ref{sec_IIB_bit}. The results show that the recommended mean mapping function is best in summary. It is also the default mapping function of the IIB descriptor. Although the quartile mapping function slightly outperforms the mean mapping function, it requires twice of descriptor size compared with the mean mapping function.

\begin{table}[tbp]
	\centering
	\scriptsize
	\begin{threeparttable}[tbp]
		\caption{\textit{mAP}, \textit{mAR}, and descriptor size of the IIB descriptors with five mapping functions $f(\bm{x})$.}			
		\label{IIB_with_mapping_rules}
		\begin{tabular}{|c|cc|cc|c|}
			\hline
			\multirow{2}{*}{Descriptor} & \multicolumn{2}{c|}{Natural dataset}                   & \multicolumn{2}{c|}{Synthetic dataset}                 & \multirow{2}{*}{\begin{tabular}[c]{@{}c@{}}Descriptor \\ size (bits)\end{tabular}} \\ \cline{2-5}
			& \multicolumn{1}{c|}{\textit{mAP}}    & \textit{mAR}    & \multicolumn{1}{c|}{\textit{mAP}}    & \textit{mAR}    &                                                                                     \\ \hline
			IIB-mean                     & \multicolumn{1}{c|}{0.9187}          & 0.8041          & \multicolumn{1}{c|}{0.9210}          & 0.7655          & 1360                                                                                \\ \hline
			IIB-max                      & \multicolumn{1}{c|}{0.8639}          & 0.7270          & \multicolumn{1}{c|}{0.8568}          & 0.6771          & 1360                                                                                \\ \hline
			IIB-min                      & \multicolumn{1}{c|}{0.8629}          & 0.7238          & \multicolumn{1}{c|}{0.8584}          & 0.6787          & 1360                                                                                \\ \hline
			IIB-quartile                 & \multicolumn{1}{c|}{\textbf{0.9270}} & \textbf{0.8139} & \multicolumn{1}{c|}{\textbf{0.9314}} & \textbf{0.7796} & 2720                                                                                \\ \hline
			IIB-sort                     & \multicolumn{1}{c|}{0.9148}          & 0.7948          & \multicolumn{1}{c|}{0.9122}          & 0.7529          & 2720                                                                                \\ \hline
		\end{tabular}
		\begin{tablenotes}
			\item The level of granularity is 4.
		\end{tablenotes}
	\end{threeparttable}
\end{table}

\subsubsection{Overlap \textit{vs.} Nonoverlap Mapping}
As mentioned in Subsection \ref{sec_IIB_bit} and Fig. \ref{fig_bit_rules} (c), every nonoverlapped four local patches in the neighborhood are used to calculate bits for the IIB descriptor by default. Using four overlapping local patches in the neighborhood can improve the performance of the IIB descriptor to an extent, as shown in TABLE \ref{IIB_with_overlaps}. However, the descriptor size will significantly increase. Therefore, nonoverlapped mapping is recommended setting for IIB descriptor formulation.

\begin{table}[tbp]
	\centering
	\scriptsize
	\begin{threeparttable}[b]
		\caption{\textit{mAP}, \textit{mAR}, and descriptor size of the IIB descriptors with or without overlap mapping.}
		\label{IIB_with_overlaps}
		\begin{tabular}{|c|cc|cc|c|}
			\hline
			\multirow{2}{*}{Descriptor} & \multicolumn{2}{c|}{Natural dataset}                   & \multicolumn{2}{c|}{Synthetic dataset}                 & \multirow{2}{*}{\begin{tabular}[c]{@{}c@{}}Descriptor \\ size (bits)\end{tabular}} \\ \cline{2-5}
			& \multicolumn{1}{c|}{\textit{mAP}}    & \textit{mAR}    & \multicolumn{1}{c|}{\textit{mAP}}    & \textit{mAR}    &                                                                                     \\ \hline
			IIB-nonoverlap               & \multicolumn{1}{c|}{0.9187}          & 0.8041          & \multicolumn{1}{c|}{0.9210}          & 0.7655          & \textbf{1360}                                                                       \\ \hline
			IIB-overlap                  & \multicolumn{1}{c|}{\textbf{0.9269}} & \textbf{0.8180} & \multicolumn{1}{c|}{\textbf{0.9326}} & \textbf{0.7872} & 4544                                                                                \\ \hline
		\end{tabular}
		\begin{tablenotes}
			\item The level of granularity is 4.
		\end{tablenotes}
	\end{threeparttable}
\end{table}

\subsubsection{Choice of Granularity}
As mentioned in Subsection \ref{sec_IIB_sampling}, the IIB descriptor is formulated in multiple granularities of feature point ROS. The IIB descriptor with a larger granularity performs better, as shown in TABLE \ref{IIB_with_granularities}. However, the descriptor size increases exponentially. To balance the descriptor size and performance, the recommended granularity for the IIB descriptor was empirically set to 4, which is the acceptable descriptor size for many visual measurement applications, such as in SLAM \cite{ORB-SLAM3}.  However, it can be adjusted according to specific visual measurement tasks in practice. 

\begin{table}[tbp]
	\centering
	\scriptsize
	\caption{\textit{mAP}, \textit{mAR}, and descriptor size of the IIB descriptors with different granularities.}
	\label{IIB_with_granularities}
	\begin{tabular}{|c|cc|cc|c|}
		\hline
		\multirow{2}{*}{Descriptor} & \multicolumn{2}{c|}{Natural dataset}                   & \multicolumn{2}{c|}{Synthetic dataset}                 & \multirow{2}{*}{\begin{tabular}[c]{@{}c@{}}Descriptor \\ size (bits)\end{tabular}} \\ \cline{2-5}
		& \multicolumn{1}{c|}{\textit{mAP}}    & \textit{mAR}    & \multicolumn{1}{c|}{\textit{mAP}}    & \textit{mAR}    &                                                                                     \\ \hline
		IIB-g2                       & \multicolumn{1}{c|}{0.5764}          & 0.3902          & \multicolumn{1}{c|}{0.5453}          & 0.3502          & \textbf{80}                                                                         \\ \hline
		IIB-g3                       & \multicolumn{1}{c|}{0.8368}          & 0.6808          & \multicolumn{1}{c|}{0.8163}          & 0.6129          & 336                                                                                 \\ \hline
		IIB-g4                       & \multicolumn{1}{c|}{0.9187}          & 0.8041          & \multicolumn{1}{c|}{0.9210}          & 0.7655          & 1360                                                                                \\ \hline
		IIB-g5                       & \multicolumn{1}{c|}{\textbf{0.9414}} & \textbf{0.8368} & \multicolumn{1}{c|}{\textbf{0.9473}} & \textbf{0.8251} & 5456                                                                                \\ \hline
	\end{tabular}
\end{table}

\subsubsection{Fast Hierarchical \textit{vs.} Brute-force Matching}

\begin{figure*}[tbp]
	\centering
	\setlength{\abovecaptionskip}{0.cm}
	\setlength{\belowcaptionskip}{-0.cm}	
	\includegraphics[width=1\textwidth]{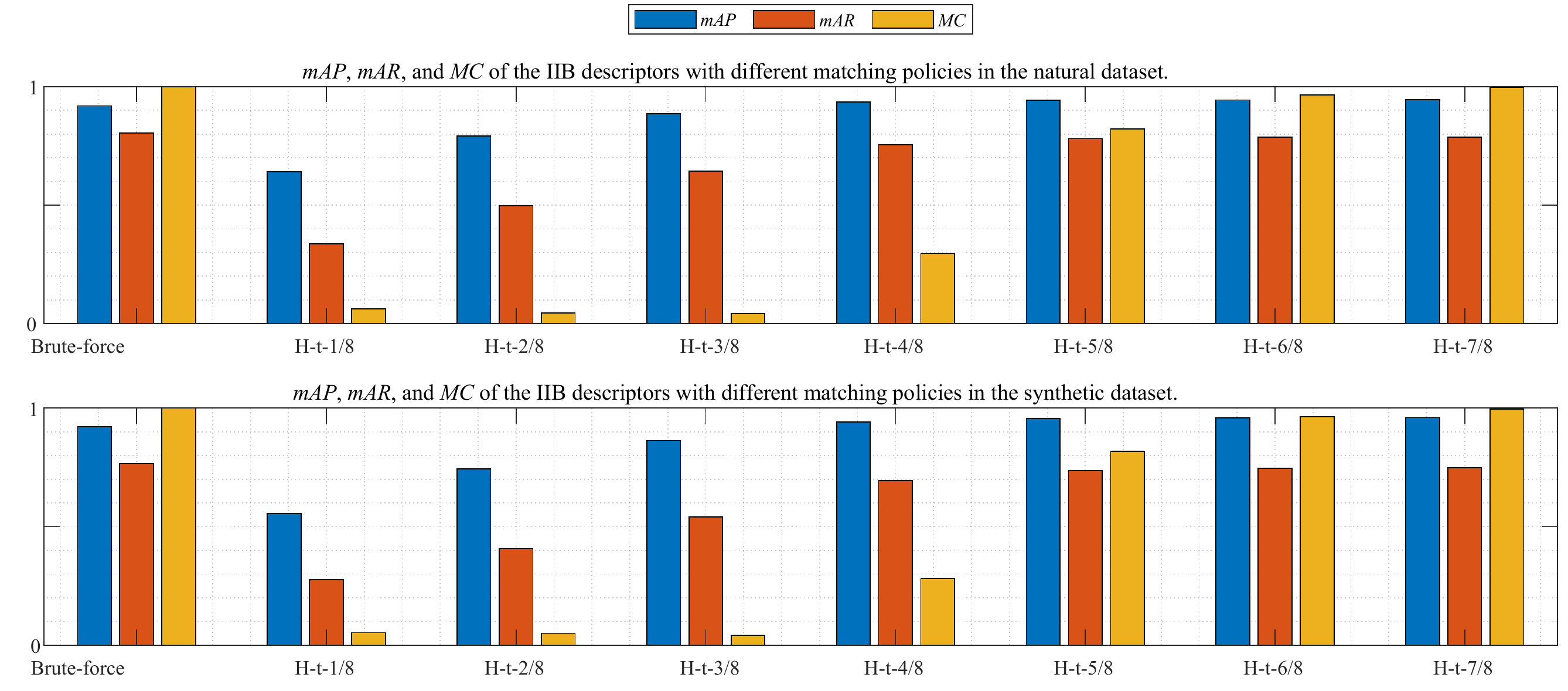}
	\caption{Performance of brute-force matching \textit{vs.} fast hierarchical matching with different thresholds for the IIB descriptor on natural and synthetic datasets, where H-t-* indicates hierarchical matching with thresholds of * $\times$ descriptor size in each granularity.}
	\label{fig_H_matching}
\end{figure*}

As mentioned in Subsection \ref{sec_hierarchical_matching}, the IIB descriptor formulation mechanism can be used to perform coarse to fine fast hierarchical matching. Fig. \ref{fig_H_matching} shows the performance and \textit{matching cost} (\textit{MC}) of brute-force and fast hierarchical matching with different thresholds for the IIB descriptor. With decreasing thresholds, the \textit{match cost} decreases. However, the corresponding performance will be degraded accordingly. In practice, a trade-off between matching cost and performance for the IIB descriptor should be attained.

\subsubsection{Reduction of Descriptor Size Using Adaboost Weights}
As mentioned in Subsection \ref{sec_adaboost_sampling}, instead of controlling relative parameters for the IIB descriptor size reduction shown in TABLE \ref{IIB_with_channels} to TABLE \ref{IIB_with_granularities}, an alternative method is to sample the quadruple of local patches to save the IIB descriptor size based on machine learning techniques. Here, the default IIB descriptors extracted from two-thirds of images in the natural dataset were used to construct the learning dataset, which is a typical binary classification dataset with a 1:1 ratio of positive to negative samples. The learning dataset was used to train the AdaBoost classifier, in which the weights of the quadruple of local patches can be learned accordingly. The learned weights for the quadruple of local patches were used to perform the sampling process for the IIB descriptor. As shown in TABLE \ref{IIB_with_adaboost}, the sampling scheme based on learned weights can significantly reduce the descriptor size while maintaining the performance of the IIB descriptor competitively.

\begin{table}[tbp]
	\centering
	\scriptsize
	\begin{threeparttable}[b]
		\caption{\textit{mAP}, \textit{mAR}, and descriptor size of the IIB descriptors with or without samplings of local patches using AdaBoost.}
		\label{IIB_with_adaboost}
		\begin{tabular}{|c|cc|cc|c|}
			\hline
			\multirow{2}{*}{Descriptor} & \multicolumn{2}{c|}{Natural dataset}                   & \multicolumn{2}{c|}{Synthetic dataset}                 & \multirow{2}{*}{\begin{tabular}[c]{@{}c@{}}Descriptor \\ size (bits)\end{tabular}} \\ \cline{2-5}
			& \multicolumn{1}{c|}{mAP}             & mAR             & \multicolumn{1}{c|}{mAP}             & mAR             &                                                                                     \\ \hline
			IIB-s128                     & \multicolumn{1}{c|}{0.7060}          & 0.5328          & \multicolumn{1}{c|}{0.6611}          & 0.4640          & \textbf{128}                                                                        \\ \hline
			IIB-s256                     & \multicolumn{1}{c|}{0.8221}          & 0.6590          & \multicolumn{1}{c|}{0.7837}          & 0.5816          & 256                                                                                 \\ \hline
			IIB-s512                     & \multicolumn{1}{c|}{0.8775}          & 0.7388          & \multicolumn{1}{c|}{0.8594}          & 0.6706          & 512                                                                                 \\ \hline
			IIB                          & \multicolumn{1}{c|}{\textbf{0.9157}} & \textbf{0.8005} & \multicolumn{1}{c|}{\textbf{0.9173}} & \textbf{0.7556} & 1360                                                                                \\ \hline
		\end{tabular}
		\begin{tablenotes}
			\item The level of granularity is 4.
		\end{tablenotes}
	\end{threeparttable}
\end{table}

\subsection{Experiments Using Predefined Feature Points with ROS}
\label{sec_exp_A}
Inspired by \cite{HPatches}, various feature points were first extracted from the reference image and then projected onto the test image according to corresponding homography matrices. Thus, every feature point in the reference image has a precise correspondence with the test image. Because these predefined feature points with ROS are the same for all descriptors, the effect of feature detectors can be excluded, and only the performance of descriptors was considered in experiments.

Here, the FAST \cite{FAST} corners were chosen to generate seed points for descriptor calculation. One thousand FAST corners were sampled uniformly in the reference image and then projected onto the test image according to corresponding homography matrices. The scale and orientation of all corners were not considered to eliminate the effect of feature scale and orientation estimation errors. Neighborhood pixels with a radius of 32 around the seed points were considered feature point ROS and used to construct feature descriptors. Nine SOTA binary descriptors, eight OpenCV-integrated and one deep learning-based descriptors, were compared in experiments. They are ORB \cite{ORB}, BRISK \cite{BRISK}, BOOST \cite{BinBoost}, BRIEF \cite{BRIEF}, LATCH \cite{LATCH}, AKAZE \cite{AKAZE}, FREAK \cite{FREAK}, BEBLID \cite{BEBLID}, and L2-Net \cite{L2-Net}. Note that, unless stated otherwise, the parameters of compared descriptors were set to default values as suggested in corresponding references or in OpenCV.

\begin{figure*}[tbp]
	\centering
	\setlength{\abovecaptionskip}{0.cm}
	\setlength{\belowcaptionskip}{-0.cm}	
	\includegraphics[width=\textwidth]{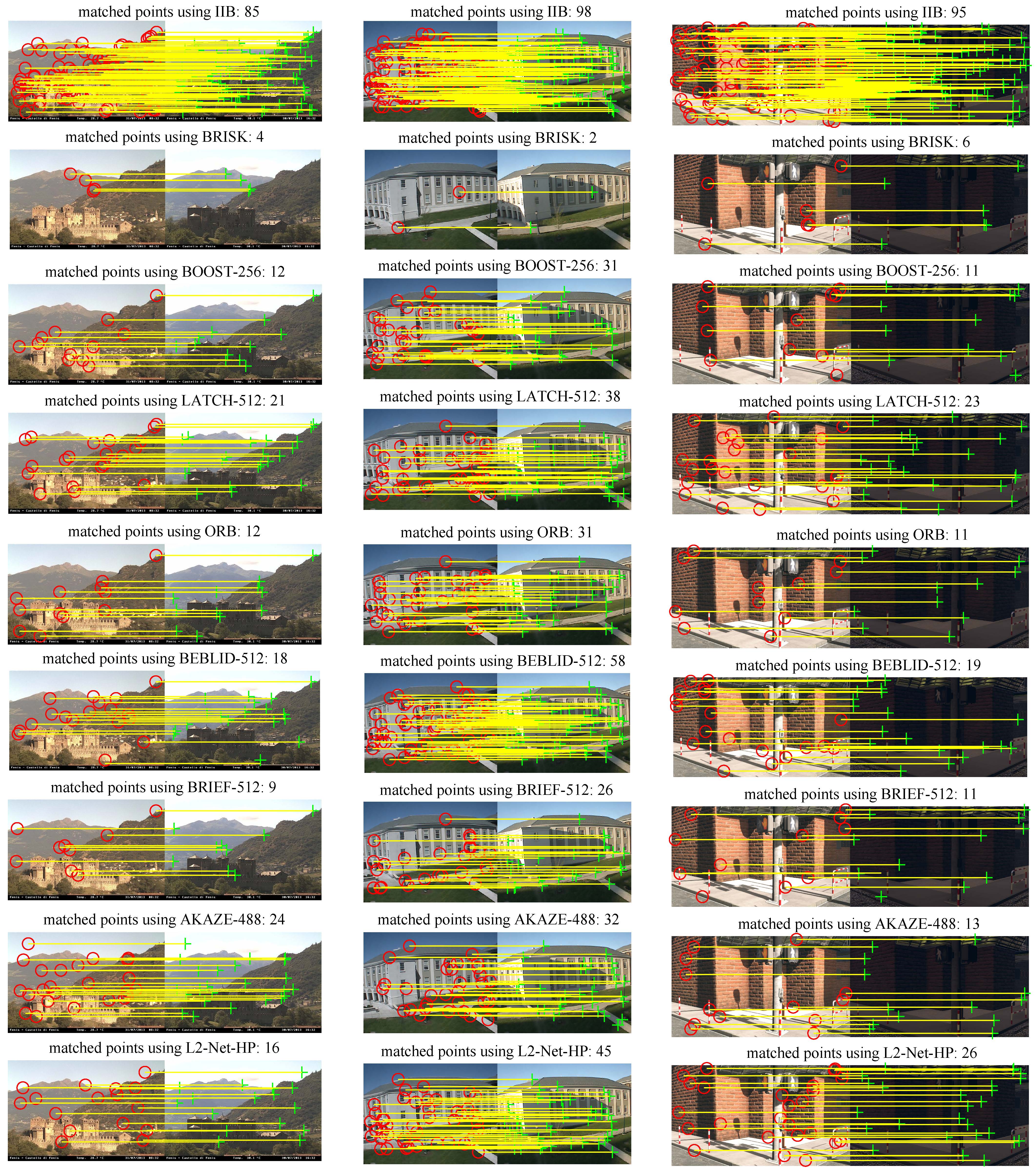}
	\caption{Matched points (sampled using a ratio of 10 for better visualization) on nature test images (the first and second columns) and synthetic test images (the third column) based on different binary descriptors.}
	\label{visual_performance}
\end{figure*}

Fig. \ref{visual_performance} visualizes the matched points using eight existing binary descriptors and the proposed IIB descriptor using mutual brute-force matching strategy. Intuitively, the inlier correspondences of the IIB descriptor are much denser than those of other descriptors, which shows the remarkable performance of the IIB descriptor qualitatively in terms of feature matching under drastic illumination variations. TABLE \ref{binary_comparsion} reports the quantized comparison results, in which IIB descriptors with different configurations are compared with nine SOTA binary descriptors. Evidently, the IIB descriptor outperformed other descriptors in terms of \textit{mAP} and \textit{mAR}, followed by the BEBLID and L2-Net descriptors. Even for the IIB-g3 and IIB-s512 descriptors with sizes similar to other descriptors, they show superiority over other SOTA descriptors.

\begin{table}[tbp]
	\centering
	\scriptsize
	\begin{threeparttable}[b]
		\caption{\textit{mAP}, \textit{mAR}, and descriptor size of four IIB descriptors \textit{vs.} various SOTA binary descriptors.}
		\label{binary_comparsion}
		\begin{tabular}{|c|cc|cc|c|}
			\hline
			& \multicolumn{2}{c|}{Natural dataset}                                               & \multicolumn{2}{c|}{Synthetic dataset}                                             &                                                                                     \\ \cline{2-5}
			\multirow{-2}{*}{Descriptor} & \multicolumn{1}{c|}{mAP}                           & mAR                           & \multicolumn{1}{c|}{mAP}                           & mAR                           & \multirow{-2}{*}{\begin{tabular}[c]{@{}c@{}}Descriptor\\ size (bits)\end{tabular}} \\ \hline
			IIB                           & \multicolumn{1}{c|}{{\color[HTML]{FE0000} 0.9187}} & {\color[HTML]{FE0000} 0.8041} & \multicolumn{1}{c|}{{\color[HTML]{FE0000} 0.9210}} & {\color[HTML]{FE0000} 0.7655} & 1360                                                                                \\ \hline
			{IIB-($\bm{\mathcal{C}}^x_g$+$\bm{\mathcal{C}}^y_g$)}                     & \multicolumn{1}{c|}{{\color[HTML]{F8A102} 0.9012}} & {\color[HTML]{F8A102} 0.7515} & \multicolumn{1}{c|}{{\color[HTML]{F8A102} 0.9008}} & {\color[HTML]{F8A102} 0.7275} & 680                                                                                 \\ \hline
			IIB-s512                      & \multicolumn{1}{c|}{{\color[HTML]{34FF34} 0.8775}} & {\color[HTML]{34FF34} 0.7388} & \multicolumn{1}{c|}{{\color[HTML]{34FF34} 0.8594}} & {\color[HTML]{34FF34} 0.6706} & 512                                                                                 \\ \hline
			IIB-g3                        & \multicolumn{1}{c|}{{\color[HTML]{34CDF9} 0.8368}} & {\color[HTML]{34CDF9} 0.6808} & \multicolumn{1}{c|}{{\color[HTML]{34CDF9} 0.8163}} & {\color[HTML]{34CDF9} 0.6129} & 336                                                                                 \\ \hline                                                                             
			ORB                           & \multicolumn{1}{c|}{0.6096}                        & 0.4193                        & \multicolumn{1}{c|}{0.5737}                        & 0.3622                        & 256                                                                                 \\ \hline
			BRISK                         & \multicolumn{1}{c|}{0.4138}                        & 0.2537                        & \multicolumn{1}{c|}{0.3708}                        & 0.2178                        & 512                                                                                 \\ \hline
			BEBLID-256                    & \multicolumn{1}{c|}{0.7697}                        & 0.5756                        & \multicolumn{1}{c|}{0.7283}                        & 0.4824                        & 256                                                                                 \\ \hline
			BEBLID-512                    & \multicolumn{1}{c|}{0.7914}                        & 0.5951                        & \multicolumn{1}{c|}{0.7514}                        & 0.4990                        & 512                                                                                 \\ \hline
			BOOST-64                      & \multicolumn{1}{c|}{0.3645}                        & 0.2184                        & \multicolumn{1}{c|}{0.3681}                        & 0.2107                        & 64                                                                                  \\ \hline
			BOOST-128                     & \multicolumn{1}{c|}{0.5375}                        & 0.3546                        & \multicolumn{1}{c|}{0.5194}                        & 0.3150                        & 128                                                                                 \\ \hline
			BOOST-256                     & \multicolumn{1}{c|}{0.6499}                        & 0.4553                        & \multicolumn{1}{c|}{0.6389}                        & 0.4054                        & 256                                                                                 \\ \hline
			BRIEF-128                     & \multicolumn{1}{c|}{0.5480}                        & 0.3664                        & \multicolumn{1}{c|}{0.5166}                        & 0.3193                        & 128                                                                                 \\ \hline
			BRIEF-256                     & \multicolumn{1}{c|}{0.5968}                        & 0.4068                        & \multicolumn{1}{c|}{0.5634}                        & 0.3505                        & 256                                                                                 \\ \hline
			BRIEF-512                     & \multicolumn{1}{c|}{0.6215}                        & 0.4272                        & \multicolumn{1}{c|}{0.5860}                        & 0.3657                        & 512                                                                                 \\ \hline
			LATCH-128                     & \multicolumn{1}{c|}{0.5878}                        & 0.4091                        & \multicolumn{1}{c|}{0.5397}                        & 0.3423                        & 128                                                                                 \\ \hline
			LATCH-256                     & \multicolumn{1}{c|}{0.6599}                        & 0.4682                        & \multicolumn{1}{c|}{0.6202}                        & 0.3956                        & 256                                                                                 \\ \hline
			LATCH-512                     & \multicolumn{1}{c|}{0.7082}                        & 0.5112                        & \multicolumn{1}{c|}{0.6748}                        & 0.4364                        & 512                                                                                 \\ \hline
			AKAZE-128                     & \multicolumn{1}{c|}{0.5698}                        & 0.4109                        & \multicolumn{1}{c|}{0.5251}                        & 0.3599                        & 128                                                                                 \\ \hline
			AKAZE-256                     & \multicolumn{1}{c|}{0.6491}                        & 0.4901                        & \multicolumn{1}{c|}{0.6043}                        & 0.4302                        & 256                                                                                 \\ \hline
			AKAZE-488                     & \multicolumn{1}{c|}{0.6922}                        & 0.5342                        & \multicolumn{1}{c|}{0.6545}                        & 0.4706                        & 488                                                                                 \\ \hline
			FREAK                         & \multicolumn{1}{c|}{0.6120}                        & 0.4216                        & \multicolumn{1}{c|}{0.4791}                        & 0.2769                        & 512                                                                                 \\ \hline
			L2-Net-HP                     & \multicolumn{1}{c|}{0.7265}                        & 0.5245                        & \multicolumn{1}{c|}{0.6890}                        & 0.4398                        & 512                                                                                 \\ \hline
			L2-Net-LIB                    & \multicolumn{1}{c|}{0.7253}                        & 0.5239                        & \multicolumn{1}{c|}{0.6963}                        & 0.4349                        & 512                                                                                 \\ \hline
			L2-Net-ND                     & \multicolumn{1}{c|}{0.7214}                        & 0.5180                        & \multicolumn{1}{c|}{0.6962}                        & 0.4334                        & 512                                                                                 \\ \hline
			L2-Net-YOS                    & \multicolumn{1}{c|}{0.7057}                        & 0.4923                        & \multicolumn{1}{c|}{0.6596}                        & 0.4054                        & 512                                                                                 \\ \hline
		\end{tabular}
		\begin{tablenotes}
			\item Colored values are the top four performance rankings.
		\end{tablenotes}
	\end{threeparttable}
\end{table}

\subsection{Experiments Combined with Feature Detectors}
\label{sec_exp_B}
Because descriptors are always used with specific feature detectors, the IIB descriptor was adapted with nine well-known feature detectors inspired by \cite{BEBLID} and compared with corresponding feature descriptors. In experiments, the detected points were the same for the IIB descriptor and compared descriptors to eliminate the influence of feature detectors. 

The nine well-known feature suits were AKAZE \cite{AKAZE}, ORB \cite{ORB}, BRISK \cite{BRISK}, SIFT \cite{SIFT}, SURF \cite{SURF}, KAZE \cite{KAZE}, SUPERPOINT \cite{SuperPoint}, ASLFEAT \cite{ASLFeat}, and ALIKE \cite{ALIKE}. The descriptors were binary for ORB, AKAZE, and BRISK, float for SURF, SIFT, and KAZE, and deep learning-based float for SUPERPOINT, ASLFEAT, and ALIKE. All descriptors were upright to eliminate the effect of feature orientation estimation errors. The parameters of nine feature suits were default values recommended by OpenCV or by the author's implementation. 

Similar to the BEBLID descriptor integrated into OpenCV, to adapt the proposed IIB descriptor to these feature detectors, the feature point ROS of the IIB descriptor was rescaled for these feature detectors. The rescaled factors were 12, 2, 4, 20, 20, and 4 for the AKAZE, ORB, BRISK, SIFT, KAZE, and SURF feature detectors, respectively. The radii of feature ROS for SUPERPOINT, ASLFEAT, and ALIKE were set to 64 pixels. The correctly matched numbers, $recall$, and $1-precision$ curves with different matching thresholds, as in \cite{BRISK}, were plotted to demonstrate the performance.

\begin{figure*}[tbp]
	\centering
	\setlength{\abovecaptionskip}{0.cm}
	\setlength{\belowcaptionskip}{-0.cm}	
	\includegraphics[width=1\textwidth]{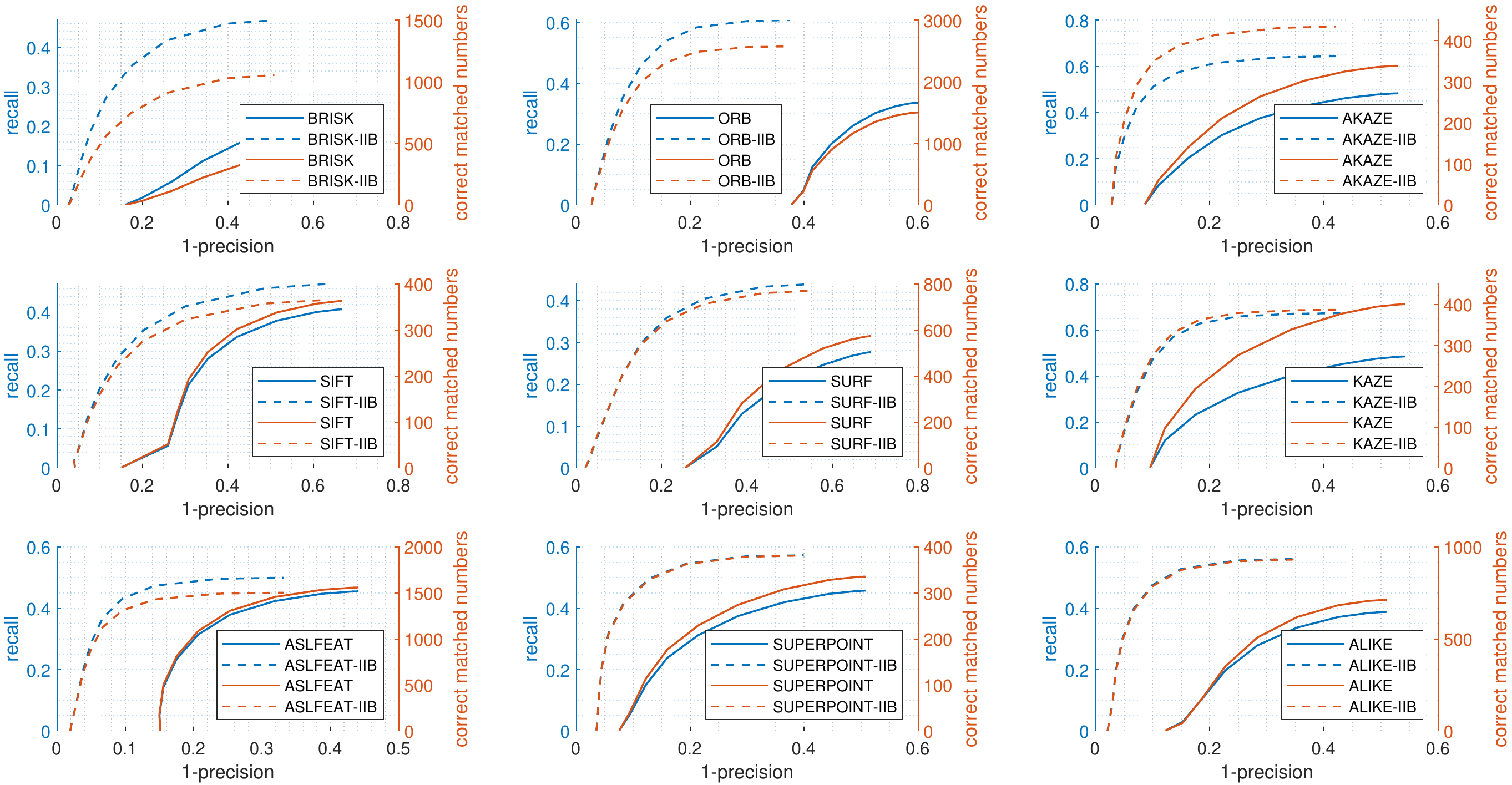}
	\caption{Using natural testing data, the correct matched numbers, recall, and 1-precision curves of the IIB descriptor with different feature suits, in which the descriptors of BRISK, ORB, and AKAZE were binary, those of SIFT, SURF, and KAZE were float, and those of ASLFEAT, SUPERPOINT, and ALIKE were deep learning-based float descriptors.}
	\label{fig_performance_with_detector_real}
\end{figure*}

\begin{figure*}[tbp]
	\centering
	\setlength{\abovecaptionskip}{0.cm}
	\setlength{\belowcaptionskip}{-0.cm}	
	\includegraphics[width=1\textwidth]{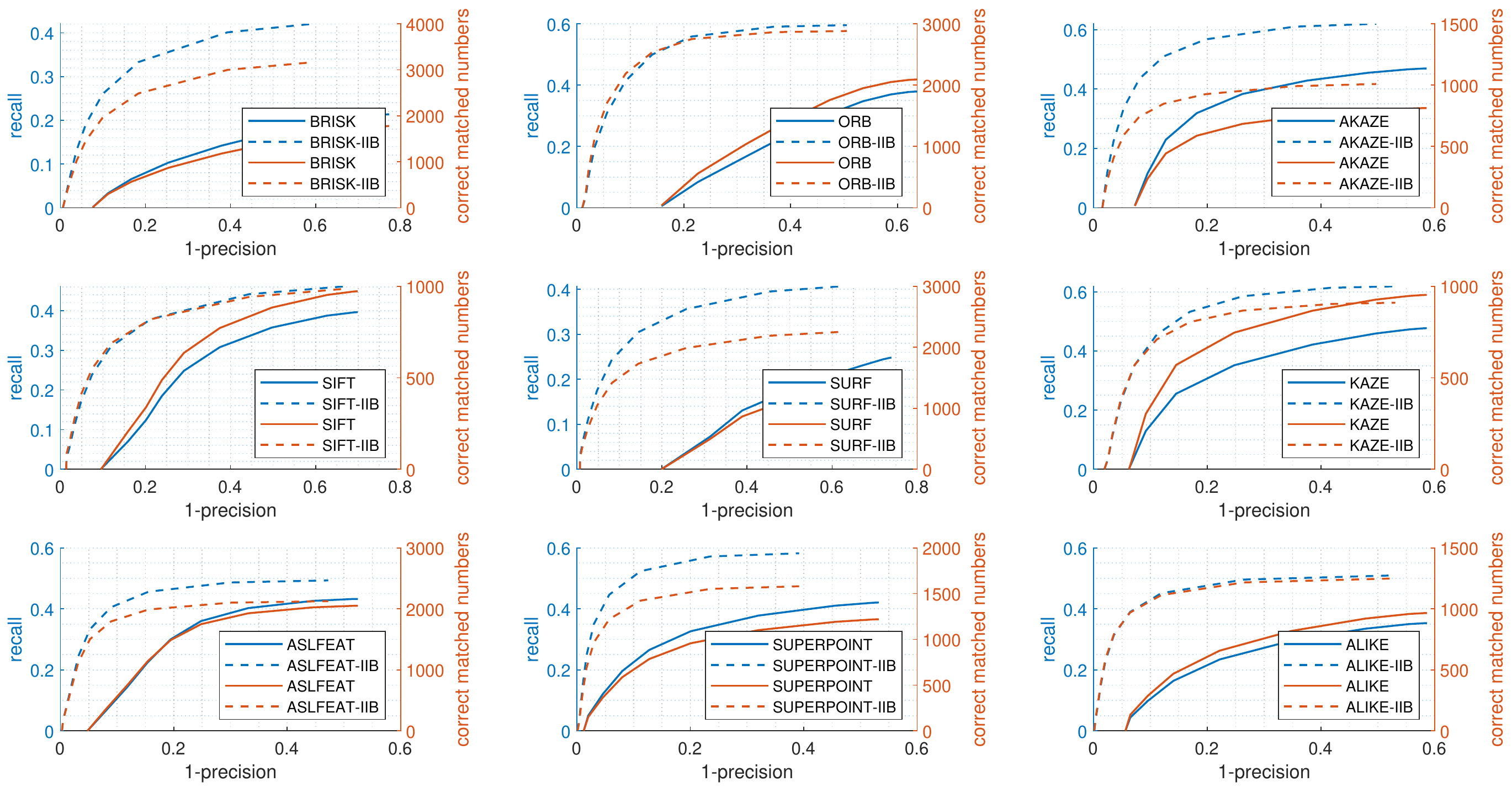}
	\caption{Using synthetic testing data, the correct matched numbers, recall, and 1-precision curves of the IIB descriptor with different feature suits, in which the descriptors of BRISK, ORB, and AKAZE were binary, those of SIFT, SURF, and KAZE were float, and those of ASLFEAT, SUPERPOINT, and ALIKE were deep learning-based float descriptors.}
	\label{fig_performance_with_detector_virtual}
\end{figure*}

Fig. \ref{fig_performance_with_detector_real} and Fig. \ref{fig_performance_with_detector_virtual} show performance comparisons of the IIB descriptors with nine feature suits on natural and synthetic datasets, respectively. Evidently, the IIB descriptor shows an overwhelming advantage over other descriptors, even for float descriptors and deep learning-based float descriptors.

\subsection{Computational Cost of the IIB Descriptor}
\label{sec_exp_C}
The computational cost for the key component of the IIB descriptor is analyzed in this subsection, referring to \cite{RAJEVENCELTHA2022101039}. Regarding the key component, \textit{i.e.}, binary descriptor bit formulation, each IIB descriptor requires $4\times M$ basic algebraic operations and $M$ relational operations, where $M$ is the intended descriptor size. The proposed IIB descriptor demonstrates competitive extraction efficiency compared to the ultrafast AKAZE/LDB descriptors, as indicated in TABLE \ref{computation_cost}. However, the memory requirement of the IIB descriptor is slightly higher than that of the memory-economic AKAZE/LDB descriptor because the IIB descriptor requires four channels of image data by default, while the AKAZE/LDB descriptor requires three instead.

In many visual measurement applications, descriptors do not require rotation invariance, rendering the computational complexity of the IIB descriptor inconsequential due to the usage of an integral image for all feature points. However, when the rotation invariance of the IIB descriptor is considered, the computational complexity increases due to re-computing the integral image for each rotated feature point ROS.

\begin{table}[tbp]
	\scriptsize
	\centering
	\caption{Computational complexity comparison between the proposed IIB descriptor and the ultrafast AKAZE/LDB descriptors, where $M$ is the intended descriptor size.}
	\begin{tabular}{|c|c|c|}
		\hline
		Method    & No. of algebraic operations & No. of relational operations \\ \hline
		AKAZE/LDB & $6\times M$                       & $M$                          \\ \hline
		IIB       & $4\times M$                       & $M$                          \\ \hline
	\end{tabular}%
	\label{computation_cost}
\end{table}

\subsection{Discussion of Experimental Results}
\label{sec_exp_analysis}
The experimental results demonstrate that the IIB descriptor outperforms other SOTA binary and float descriptors in the designed experiments. Moreover, it offers competitive computational efficiency, even compared to the ultrafast AKAZE/LDB descriptors. The superior performance of the IIB descriptor is due to its ability to encode the illumination-invariant characteristic effectively. Specifically, the quadruple inter-patch invariance among local patches in multiple spatial granularities is a key factor to achieve this feature.

It has been reported that deep learning-based descriptors have shown powerful feature representation ability. In addition to comparing the illumination-insensitive ability of descriptors, the IIB descriptor offers three additional key merits over deep learning-based descriptors. (1) Deep learning-based methods require abundant and carefully labeled data covering images under different lighting conditions that are difficult to obtain in practice. Instead, the proposed IIB descriptor has no requirement for such training data and still achieves excellent illumination insensitivity, making the IIB descriptor an attractive option for scenarios where acquiring labeled data is difficult or impractical. (2) Deep learning-based methods usually require powerful hardware configurations and GPU devices. Instead, the proposed IIB descriptor can be effectively applied in resource-constrained systems and platforms. (3) Deep learning-based methods typically require extensive computational resources and are challenging to execute in real time, even with the assistance of GPU devices. Instead, the efficiency of the IIB descriptor is competitive with the ultrafast AKAZE/LDB descriptors due to the usage of integral images. The merits of (2) and (3) make the IIB descriptor a practical and versatile solution for a wide range of applications, especially for those with limited resources.

\section{Application for Long-term Visual Localization}
\label{sec_app}
In this study, the proposed IIB descriptor was used to perform a typical visual measurement task, \textit{i.e.}, long-term visual localization in \cite{our_demo} based on the Ford AV dataset\footnote{https://avdata.ford.com/} (route "Log 5"). The major challenge is the drastic illumination variations of image data. The dataset included seasonal images of the same roads recorded in different months. A visual localization database was created using images recorded in August. Visual localization was performed using images recorded in October by matching image features with those in the pre-built visual localization database.

In the visual localization system, the ORB\cite{ORB} and AKAZE\cite{AKAZE} point features with IIB descriptors and the line feature \cite{LSD} with descriptors \cite{LBD} were used to perform robust matching between the test image and the pre-built visualization localization database. The localization system with the IIB descriptor has a centimeter-level positioning accuracy and high orientation accuracy. This is a remarkable performance compared with other SOTA methods. TABLE \ref{results_localization_results} lists some examples, in which IMU, GNSS, and WSS are the inertial measurement unit\footnote{https://en.wikipedia.org/wiki/Inertial\_measurement\_unit}, global navigation satellite system\footnote{https://en.wikipedia.org/wiki/Satellite\_navigation}, and wheel speed sensor\footnote{https://en.wikipedia.org/wiki/Wheel\_speed\_sensor}, respectively. The supplementary material provides a video demonstration of the localization system using the IIB descriptor. 

\begin{table}[tbp]
	\centering
	\scriptsize
	\setlength\tabcolsep{3pt}
	\caption{Reported performance comparison of eight camera-based localization systems in their studies.}
	\label{results_localization_results}
	\begin{threeparttable}[b]
		\begin{tabular}{|c|c|ccc|}
			\hline
			\multirow{2}{*}{Method} & \multirow{2}{*}{Inputs}    & \multicolumn{3}{c|}{Positioning Accuracy   (Mean)}                                       \\ \cline{3-5} 
			&                            & \multicolumn{1}{c|}{Lateral}      & \multicolumn{1}{c|}{Longitudinal} & Altitude         \\ \hline
			\textbf{Proposed}              & \textbf{Camera, IMU,   DB} & \multicolumn{1}{c|}{\textbf{5cm}} & \multicolumn{1}{c|}{\textbf{5cm}} & \textbf{2cm}     \\ \hline
			{\cite{9784872_2}}                  & Camera, DB                 & \multicolumn{1}{c|}{9cm}          & \multicolumn{1}{c|}{19cm}         & 15cm             \\ \hline
			{\cite{9772400_2}}                  & Camera, GNSS, WSS, IMU, DB & \multicolumn{1}{c|}{4cm}          & \multicolumn{1}{c|}{17cm}         & \textbackslash{} \\ \hline
			{\cite{s22072434}}               & Camera, GNSS, DB           & \multicolumn{2}{c|}{30cm (two-dimensional)}                                & \textbackslash{} \\ \hline			
			{\cite{9561459}}                  & Camera, WSS, GNSS, DB      & \multicolumn{1}{c|}{12cm}         & \multicolumn{1}{c|}{43cm}         & \textbackslash{} \\ \hline
			{\cite{9341003}}                  & Camera, WSS, DB            & \multicolumn{1}{c|}{7cm}          & \multicolumn{1}{c|}{73cm}         & 6cm              \\ \hline			
			{\cite{7547970}}                  & Camera, GNSS, WSS, IMU, DB & \multicolumn{1}{c|}{58cm}         & \multicolumn{1}{c|}{143cm}        & \textbackslash{} \\ \hline						
			{\cite{7313499}}                  & Camera, GNSS, IMU, DB      & \multicolumn{3}{c|}{73cm (three-dimensional)}                                            \\ \hline
		\end{tabular}
		\begin{tablenotes}
			\item - IMU means inertial measurement unit.
			\item - GNSS means global navigation satellite system.
			\item - WSS means wheel speed sensor.
		\end{tablenotes}
	\end{threeparttable}
\end{table}

\section{Conclusion and Future Work}
\label{sec_conclusion}
This study presented an illumination-insensitive descriptor to address the feature point matching problem under drastic illumination variations in the view of feature description. The local inter-patch invariance exhibited in multiple spatial granularities of feature point ROS was fully exploited to encode feature points from coarse to fine, thus facilitating computationally efficient hierarchical matching. Except for pixel intensity and related gradient data of the gray image, the proposed IIB descriptor can be extended by using additional illumination-insensitive image data, such as depth, texture, and semantic segmentation maps from other sensors and algorithms. The IIB descriptor has an inconsequential computational cost owing to the usage of integral images. Numerical experiments show that the IIB descriptor performs remarkably compared to SOTA binary descriptors and some testing float descriptors and can be successfully employed in a long-term visual localization system.	

This study aims to design a highly efficient and illumination-insensitive binary descriptor. Although the IIB descriptor exhibits impressive performance in illumination variance scenarios, it has a major limitation due to the omission of time-consuming processes, \textit{e.g.}, feature point ROS normalization, to ensure excellent extraction efficiency. These processes are essential for achieving viewpoint invariance. As a result, the IIB descriptor may not perform optimally in scenarios with significant viewpoint variances, \textit{e.g.}, wide baseline image matching \cite{10.1007/s11263-020-01385-0}. To address this, incorporating viewpoint invariances through these processes could be explored in the future, provided that efficiency requirements are met.

\section*{Acknowledgments}
This research is partially supported by the National Natural Science Foundation of China (NSFC, No. U19A2052, 62171088, and 62171302), in part by the Sichuan Science and Technology Program under Grant No. 2023NSFSC1965 and the 111 Project under Grant No. B21044, and by the EU Horizon 2020 program towards the 6G BRAINS project H2020-ICT 101017226.

\bibliographystyle{IEEEtran}

\end{document}